\documentclass[10pt,twocolumn,romanappendices]{IEEEtran}

\usepackage[noadjust]{cite}
\usepackage[numbers,square,comma,sort&compress]{natbib}
\usepackage{calc}
\usepackage{comment}
\usepackage{graphicx}
\usepackage{amsmath,epsfig,amssymb,verbatim,amsopn,cite,subfigure,multirow}
\usepackage{balance}
\usepackage[usenames,dvipsnames]{color}
\usepackage[all]{xy}  
\usepackage{url}
\usepackage{amsfonts}
\usepackage{amssymb}
\usepackage{epsfig}
\usepackage{slashbox}
\usepackage{bm}
\usepackage{epsfig}
\usepackage{amsmath,amssymb, amstext}
\usepackage{graphicx}
\usepackage{epstopdf}
\usepackage{algorithm}
\usepackage{algorithmic}
\usepackage[numbers,square,comma,sort&compress]{natbib}
\allowdisplaybreaks

\begin{document}
\title{\huge A Framework for Super-Resolution of Scalable Video via Sparse Reconstruction of Residual Frames}
\author{Mohammad Hossein~Moghaddam$^\dag   $, Mohammad Javad~Azizipour$^\dag $, Saeed~Vahidian$^* $,\\ and Besma~Smida$^*$,~\IEEEmembership{Senior Member,~IEEE}\\
\small
$^\dag $Department of Electrical Engineering, K.N. Toosi University of Technology, Tehran, Iran\\
Email: \{mhmoghadam, mj.azizipour\ @ee.kntu.ac.ir\} \\
$^*$Department of Electrical and Computer Engineering, University of Illinois at Chicago (UIC), Chicago, IL, USA\\

Email: \{svahid2, smida\ @uic.edu\} \\
}

\normalsize



\maketitle

\begin{abstract}
This paper introduces a framework for super-resolution of scalable video based on compressive sensing and sparse representation of residual frames in reconnaissance and surveillance applications. We exploit efficient compressive sampling and sparse reconstruction algorithms to super-resolve the video sequence with respect to different compression rates. We use the sparsity of residual information in residual frames as the key point in devising our framework. Moreover, a controlling factor as the compressibility threshold to control the complexity-performance trade-off is defined. Numerical experiments confirm the efficiency of the proposed framework in terms of the compression rate as well as the quality  of reconstructed video sequence in terms of PSNR measure. The framework leads to a more efficient compression rate and higher video quality compared to other state-of-the-art algorithms considering performance-complexity trade-offs.
\end{abstract}

\section{Introduction}
The growth of video traffic for live and on-demand content has been a challenging demand in reconnaissance and surveillance applications to provide video streaming with  high quality viewing, low bandwidth consumption and low processing complexity, considering video compression algorithms \cite{bennett2005operational,bennett2008scalable,bhaskaranand2011low,ullah2011energy}. 
The general solutions for video compression are proposed in scalable video coding (SVC) \cite{draft2003recommendation} and the more recent scalable high efficiency video coding (SHVC) \cite {sullivan2013standardized} that has been studied extensively over the last several decades and is now supported in several video codecs and standards such as H.264/SVC and H.265/HEVC.

SVC includes several scalability methods including temporal, spatial and quality (or SNR) scalabilities \cite{schwarz2007overview}. These compression schemes typically target multimedia applications such as video storage and playback, video streaming, video conferencing, and entertainment broadcasting, but reconnaissance and surveillance applications need video compression requirements different from the typical video compression applications. In most of reconnaissance and surveillance applications, for example, for unmanned aerial vehicles (UAVs) and wireless video sensor networks (WVSNs), required bandwidth, power, storage capacity and processing power are the concerning challenges for video streaming. One of the most popular methods among emerging compression and reconstruction methods is the concept of compressive sensing (CS). CS has received attention in fields such as image and video processing \cite{R12}. In the wake of extensive advances in CS, several video processing algorithms are proposed based on the concept of sparse representation of video sequence in various domains such as Discrete Fourier Transform (DFT), Discrete Wavelet Transform (DWT), Contourlet Transform (CLT) \cite{li2013new} and some new state-of-the-art transforms like STone transform \cite{goldstein2015stone}. Additionally, some interesting solutions are made using CS in conjunction with SVC \cite{jiang2012scalable,li2013new,wang2015lisens}. Several efforts are also made for super-resolving  scalable video considering bandwidth efficiency and quality enhancement \cite{kulkarni2012understanding,chen2012video}. Super-resolution (SR) performance could be boosted by exploiting CS as a compression/reconstruction method in cases of bandwidth reduction scenarios that is known under the name of "compressive sensing super-resolution" \cite{yang2010image,mahfoodh2015super}.

High compression rate, high video quality, and low computational complexity are the desired factors in video streaming which generally could not be achieved simultaneously and there has been always some tradeoffs between them. According to the aforementioned explanations, in this paper we aimed to propose proper solutions for efficient video streaming by utilizing compressive sensing concepts in scalable video super-resolution which can be used for reconnaissance and surveillance applications according to restricted conditions on bandwidth requirements.
The fundamental goal of super-resolution algorithms is to retain the original high-resolution content from low-resolution ones to have desirable compression gains. Video streaming standards need techniques that stream video sequence with acceptable quality at lower data rates.
There are several state-of-the-art algorithms for video compression in the literature \cite{jiang2012scalable,mahfoodh2015super,mukherjee2013latest}. In some of them, the sparsity of frames in the frequency domain is utilized to achieve compression gain, such as using FT to represent the signal in frequency domain as a sparse signal \cite{jiang2012scalable}. In some other approaches, motion vectors and some state-of-the-art methods like quad-tree sectorization are utilized in time domain to achieve compression gain, such as the algorithms used in VP9 codec proposed by Google Corp. \cite{mahfoodh2015super,mukherjee2013latest}.

In this paper, as an extension to our previous work \cite{moghaddam2015performance}, we propose a framework based on sparse representation of multiple residual layers of scalable video for compressive super-resolving of standard video frames in the time domain. We have considered the fact that the residual information embedded in residual frame which is constructed from difference of original and up-sampled frames, has sparse nature and could be considered for compressive sampling and efficient sparse reconstruction. In our framework, we have defined a compressibility threshold (CT) for residual information in multiple layers of scalable video sequence that is used for controlling the complexity of reconstruction algorithms considering performance-complexity tradeoffs. \\

The advantages of our proposed framework with current state-of-the-art algorithms are threefold: $i$) Our framework works in the time domain and does not suffer from high computational complexity of changing domains from time to frequency and reverting back (as in algorithms which utilize the DFT). $ii$) Our proposed framework does not need to utilize any motion estimation algorithm which again decreases the load of computational complexity. $iii$) Our framework proposes better compression rates in comparison with current state-of-the-art algorithms such as motion vector estimation with quad-tree sectorization \cite{mahfoodh2015super}. We also considered the complexity of sparse reconstruction methods, and defined the CT to make our framework flexible toward different performance-complexity tradeoffs.

The remainder of this paper is organized as follows, Section II, explains in detail problem formulation and proposed framework. In Section III, we evaluate the performance of the proposed framework by numerical experiments and finally, Section IV
concludes the paper.

\section{Problem Formulation and Proposed Method}
\subsection{Compressive Sensing review}
Compressive sampling was first introduced in \cite{R11,R12}. It resulted from applying the sparsity condition to sampling. The first motivation for employing CS technique is the sparsity of the signal in certain domains (time, frequency, wavelet, etc.). Fortunately, this condition is met in most engineering applications including image and video processing. Let $\mathbf{x} \in {\mathbb{R}}^n$ be a vector that represents discrete and finite signal in time domain. We would consider $\mathbf{x}$ as a discrete $k$-sparse signal if its projection onto an orthonormal space, included at most $k$ nonzero components. Mathematically it can be written as
\begin{equation}
\mathbf{x} = \mathbf{\Psi} \mathbf{s},
\label{E5}
\end{equation}
where $\mathbf{\Psi} \in \mathbb{R}^n$ is a unitary matrix and $\mathbf{s} \in \mathbb{R}^n$ is a vector with at most $k$ nonzero components. In compressive sampling, a linear and non-adaptive sampling is preferred as:
\begin{equation}
\mathbf{y} = \mathbf{\Phi} \mathbf{x} = \mathbf{A} \mathbf{s},
\label{E6}
\end{equation}
where $\mathbf{A} = \mathbf{\Phi} \mathbf{\Psi}$ is an $M \times N$ matrix, independent of $\mathbf{s}$ and known as sensing matrix. The dimension of the sensing matrix defines the compression rate in the sense that an $N$-dimensional vector is transformed to an equivalent $M$-dimensional vector ($M \ll N$). The objective of CS is to reconstruct $\mathbf{x}$, or the equivalent sparse vector $\mathbf{s}$. If we assume Eq. \eqref{E6} and $M \ll N$, we get an under-determined system of linear equations with infinitely many solutions, but by having sparsity as a necessary condition, we may achieve a unique solution. Many methods are introduced to solve the above problem. In this work,  we have utilized the most  prominent algorithms: $\ell_1$ minimization, Orthogonal Matching Pursuit (OMP) \cite{tropp2007signal} and Compressed Sampling Matching Pursuit (CoSaMP) \cite{needell2009cosamp}. $\ell_1$ minimization is introduced as in the following:
\begin{equation}
\mathbf{\hat{s}} = \mathrm{argmin} \| \mathbf{s} \|_{1},  \quad s.t.\ \mathbf{y} = \mathbf{A} \mathbf{s}.
\label{E7}
\end{equation}
In \cite{R12}, the author demonstrated that the solution of (3) is very close to the optimal solution in theory. In addition, if the sample vector is noisy, the problem can be re-written as
\begin{equation}
\mathbf{\hat{s}} = \mathrm{argmin} \| \mathbf{s} \|_{1},  \quad s.t. \ \|\mathbf{y}- \mathbf{A} \mathbf{s}  \|_2 < \sigma,
\label{E8}
\end{equation}
where $\sigma$ is proportional to the noise variance as $\| n\|_{2}^{2} \leq \sigma^{2}$. OMP and CoSaMP are different extensions of the same family known as greedy pursuit family \cite{tropp2007signal,needell2009cosamp}.
The OMP algorithm, finds a column $\boldsymbol \phi_i$ which has the highest correlation with residual vector $r$ at each iteration. The residual vector is equal to measurement vector $y$ for the first iteration. The index of this column that indicates the location of nonzero element gathers into $\Lambda$ set \cite{tropp2007signal}. Then, the amplitude of nonzero element is calculated by least square (LS) problem, i.e. $\mathop {{\rm{argmin}}}\limits_x {\left\| {y - \Phi x} \right\|_2}$ . Finally, the effect of the aforementioned column will be eliminated by using orthogonal projection, $r = y-\Phi_{\Lambda} x$ and the process will be repeated with new residuals. Another algorithm in greedy pursuits family is CoSaMP which proposed in \cite{needell2009cosamp}. This approach is based on OMP, but it has been shown that the CoSaMP algorithm has tighter bounds on performance and convergence. Each iteration of this algorithm includes five major steps: identification, support merger, estimation, pruning and sample update. The number of iterations can be determined by a halting criterion such as sparsity order and desired error norm \cite{needell2009cosamp}.

\subsection{Scalable Video Super-Resolution Approach}
Considering spatial scalability, the quality and performance of multi-layer video streaming is enhanced using CS and sparse reconstruction methods. Different layers in the multi-layer video sequence can be decoded with a reasonable complexity. They are used to reconstruct the original video sequence at the receiver side, with different video qualities. The 2-layers scalable video structure for the original raw video sequence $\mathbf{V}$ is shown in Fig. 1.
\begin{figure}
\centering
\includegraphics[width=9cm]{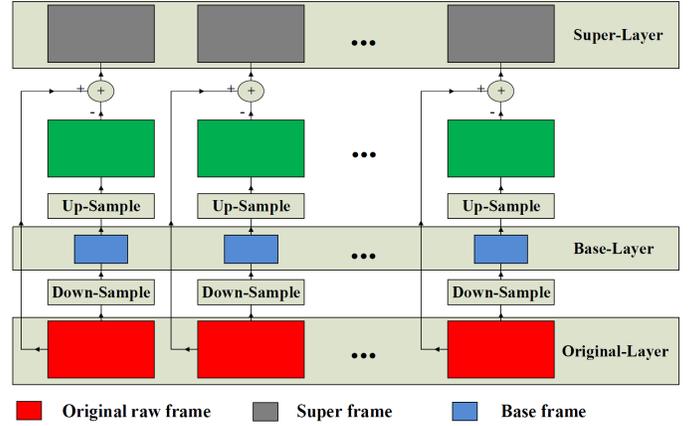}
  \caption{Spatial Scalable Video Structure}
  \vspace{-5mm}
\label{SOP_SNR}
\end{figure}
The structure contains one base-layer that includes down-sampled video frames, and one super-layer (or more super-layers in case of multi-layer scalable video) that contains residual video frames. The super layer is obtained via spatial down-sampling and up-sampling of the original raw frames as follows:

\begin{equation}
\mathbf{V}_{\rm{sup} }^{\emph{j}} = {\mathbf{V}^{\emph{j}}} - \mathbf{V}_{\rm{up}}^{\emph{j}},
\end{equation}
\begin{equation}
\mathbf{V}_{{\rm{up}}}^{\emph{j}}\, = \,{\mathbf{U}_\emph{m}}.\mathbf{V}_{\rm{down}}^{\emph{j}},
\end{equation}
\begin{equation}
\mathbf{V}_{\rm{down}}^{\emph{j}}\, = \,{\mathbf{D}_\emph{m}}.{\mathbf{V}^{\emph{j}}},
\end{equation}

\noindent where $\mathbf{V}_{\sup }^{j}$ indicates the super-frame $j$, and $\mathbf{V}_{down}^{j}$ and $\mathbf{V}_{up}^{j}$ indicate down-sampled and up-sampled video frames. $\mathbf{U}_m$ and $\mathbf{D}_m$ indicate up-sampling and down-sampling filters with rate $m$, and $\mathbf{V}^{j}$ indicates the original raw video frame $j$. At the receiver side, the reconstruction procedure without using the super-layer is straightforward as up-sampling of the received video sequence; while simply just up-sampling the video sequence gives us a rough
blurry version of the original sequence. But it could be handled by super-resolving the video frames using residual information provided in super-layers. The super-resolved video frame $\mathbf{Y}^{j}$ can be illustrated as:

\begin{equation}
{\mathbf{Y}^{\emph{j}}} = \,{\mathbf{U}_\emph{m}}.\mathbf{V}_{\rm{down}}^{\emph{j}}\, + \,\mathbf{V}_{\rm{sup} }^{\emph{j}},
\end{equation}
\begin{equation}
\mathbf{V}_{\rm{down}}^{\emph{j}}\, = \,{\mathbf{D}_\emph{m}}.{\mathbf{V}^{\emph{j}}},
\end{equation}

\noindent and after the super-resolution step, the final super-resolved video frame would be produced as $\mathbf{Y}^{j}$.
\subsection{Proposed Compressive Super-Resolution Approach }
We proposed a model based on compressive sampling and sparse reconstruction, illustrated in Fig. 2.
The residual frames, in all standard video sequences have many pixel values that are very small compared to  the nominal peak value. We defined a CT and for all pixel values which are lower than the CT value, we set them to zero. The resulting signal is a sparse signal that can be  compressed  at a desirable compression rate and low complexity using CS methods.  By modeling the residual frame as a signal with sparse representation, we prove that it is compressible. This approach could leverage the complexity of sparse reconstruction algorithms, since it would decrease the number of iterations in inner loops according to tighter search domain. The compression rate is defined as $M/N$ where $M$ and $N$ are the number of rows and columns of a Gaussian sampling matrix. It can be shown that with the following number of measurements
\begin{figure}
\centering
\includegraphics[width=9cm]{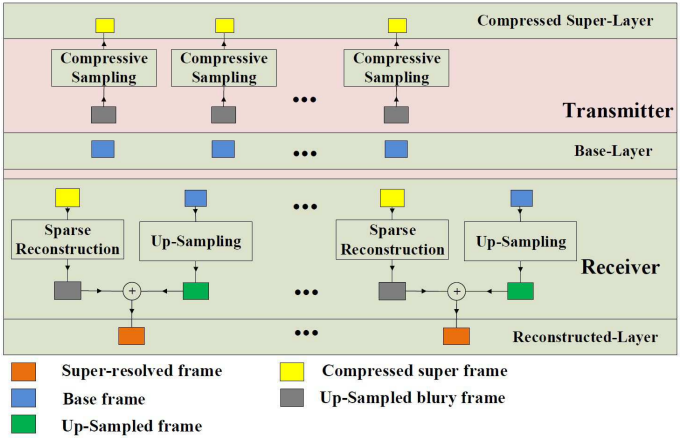}
\vspace{-7 cm}
\caption{Proposed framework}
\label{SOP_SNR}
\vspace{-5mm}
\end{figure}

\begin{equation}
M = \mathcal{O}(k\log (N/k)),
\end{equation}
the sampling matrix could satisfy the restricted isometry property (RIP) condition with a high probability, and we can recover any $k$-sparse signal perfectly \cite{eldar2012compressed}. Compressive sampling of residual frames in transmitter can be presented as:
\begin{equation}
\mathbf{v}_{\rm{sup \_CS}}^{\emph{j}}(:) = {\mathbf{\Psi} _{\emph{M} \times \emph{N}}}({\Omega _{\rm{CT}}} \odot \mathbf{v}_{\rm{sup} }^{\emph{j}}(:)),
\end{equation}
\noindent where ${\mathbf{\Psi} _{M \times N}}$ is a Gaussian random matrix with compression rate of $M/N$, $\mathbf{v}_{\sup }^{j}(:)$ is the vectorized version of $\mathbf{V}_{\sup }^{j}$, ${\mathbf{\Omega} _{ct}}$ indicates the zero-forcing vector defined according to a pre-defined threshold CT, Symbol $\odot$ indicates element-wise multiplication, and ${\mathbf{\Omega} _{CT}}$ is a binary vector to the size of $\mathbf{v}_{\sup }^{j}(:)$ which has zeros on matrix elements which are lower than the CT and ones in other places. By executing ${\Omega _{ct}} \odot \mathbf{v}_{\sup }^{j}(:)$, the
compressible vector $\mathbf{v}_{\sup }^{j}(:)$ changes to an sparse vector and after compressive sampling by Gaussian matrix ${\mathbf{\Psi} _{M \times N}}$, the resulting compressed frame is constructed and placed as a part of compressed super-frame.
At the receiver side, for sparse reconstruction using $\ell_1$ minimization and according to Eqs. (1-3) and (11) we have
\begin{equation}
\hat{\textbf{s}}^{\emph{j}}_{\rm{sup}}(:) = \mathrm{argmin} \| \mathbf{s}^{\emph{j}}_{\rm{sup}}(:)  \|_{1}, \quad s.t. \ \mathbf{v}^{\emph{j}}_{\rm{sup-CS}} (:) =  \mathbf{A} \mathbf{s}^{\emph{j}}_{\rm{sup}}(:).
\end{equation}
After sparse reconstruction of super-frame vectors $\mathbf{v}^{j}_{\rm{sup-CS}} (:)$, the super-resolved frames can be found as
\begin{equation}
\mathbf{V}^{\emph{j}} = \mathbf{V}^{\emph{j}}_{\rm{sup-CS}} + \mathbf{V}_{\rm{up}}^{\emph{j}},
\end{equation}
where $\mathbf{V}_{up}^{j}$ is the up-sampled version of the base-frame $j$. It is important to note that the choice of the pair $(M/N,\rm{CT})$  affects the performance and the complexity of proposed framework. One efficient way of choosing the CT is to scan the energy of compressible signal and set the value of the CT according to the peak value of all pixels comparing with other pixel values. But since the CT and $M/N$ are inter-related,  we should decide on them jointly. For example, as we found empirically, one could set the number of measurements ($M$), and then decide on the CT as a pixel value which after zero-forcing by ${\Omega _{ct}}$, make the sparsity level of ${\mathbf{\Omega} _{ct}} \odot \mathbf{v}_{\sup }^{j}(:)$ to one third or one fourth of the number of measurements, or even less (not higher than one third). Deciding on $M/N$ is not trivial, several bounds are proposed based on various theorems which can tighten the range of $M/N$ such as the bounds proposed according to Walsh Theorem  \cite{R11} or the bounds which are proposed based on coherency of sampling matrix \cite{eldar2012compressed}, that are out of the scope of this paper. It is just empirically found that three or four times of the order of sparsity could be a good choice for $M$. Further approaches for deciding on pair $(M/N,\rm{CT})$ according to an optimization method and its efficient solutions are left for future works.
\vspace{-2mm}
\begin{figure}
  \centering
\includegraphics[width=3.2 in]{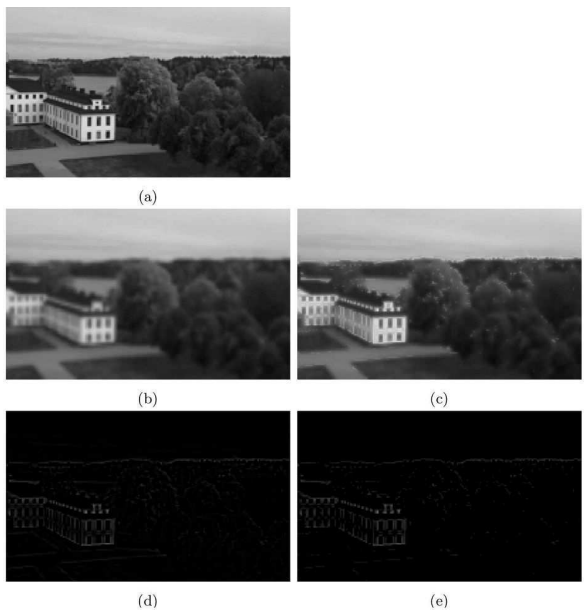}
  \vspace{-3.5cm}
  \caption{Simulation results for frame 1 of "in to the tree" video sequence, using OMP method with $M/N$ = 0.3 and CT=15: (a) original frame, (b) up-sampled frame of base frame, (c) reconstructed frame based on proposed model, (d) residual super-frame before compressive sampling, and (e) reconstructed residual super-frame after sparse reconstruction.}
\end{figure}

\vspace{-2mm}
\begin{figure}
  \centering
\includegraphics[width=9cm]{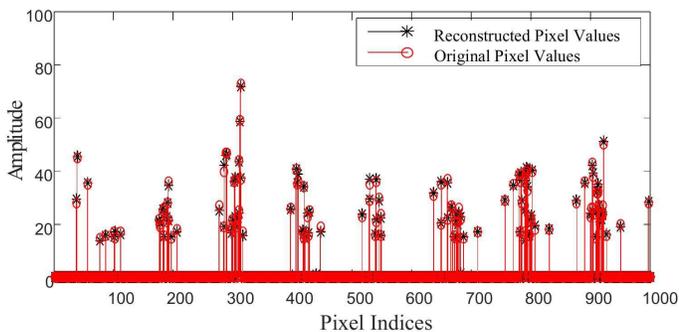}
  \vspace{-8cm}
  \caption{Simulation results for frame 1 of "in to the tree" video sequence, using OMP method with $M/N$ = 0.3 and CT=15: actual pixel values in Fig3-(d) and corresponding reconstructed ones for 1000 pixels of residual super-frame.}
  \vspace{-5mm}
\end{figure}

\vspace{+5mm}
\section{Numerical Experiments}

We used "in to the tree" standard video sequence as the original raw video sequence to compare our results with those in \cite{mahfoodh2015super} as the most recent work in general video compression schemes. Three reconstruction algorithms i.e., OMP, CoSaMP, and $\ell_1$-minimization are exploited for the sparse reconstruction of compressive-sampled super-frames. Fig. 3 sketches the simulation results for frame 1 of "in to the tree" video sequence. Fig. 4 illustrates the performance of reconstruction algorithm by indicating actual pixel values in Fig. 3-(d) and corresponding reconstructed ones for 1000 pixels of residual super-frame. Fig. 5 depicts the performance of the proposed framework versus the CT. As is evident form the figure, for the CT values more than 35, the three methods have similar performances which is due to the fact that according to the CT, excessive number of pixel values are omitted, i.e., by increasing the CT, we lose more data. Fig. 6 presents the performance of the proposed framework versus the compression rate. We observe that, by increasing the compression rate, an upper error floor happens after a certain point and slight changes in performance of all three methods take place. It can be explained in this way that, by increasing the compression rate up to a certain rate (here $M/N=0.3$) the compressive sampling can track most of the sparse points and reconstruct the sparse signal to the best point.

Turning our attention to Fig. 7 where we compared the performance of the proposed framework with that of the motion vector estimation with quad-tree sectorization algorithm (we call it quad tree algorithm for short) proposed in \cite{mahfoodh2015super}. Simulation result reveals that, the performance of our proposed framework outperforms that of the quad tree algorithm in the testbench of PSNR for each single frame. The performance of proposed framework is also constant in comparison with decreasing performance of quad tree algorithm, as the number of frames grows higher. Constant values of PSNR is due to the structure of the proposed framework which defines a super-frame for each base-frame which yields constant performance for all frames.

Additionally, in what follows, some interesting observations are drawn from Figs. 5 to 7. $(i)$ In quad tree algorithm \cite{mahfoodh2015super}, there is only one super-frame for each group of pictures (GOP) which leads to some degradations in performance as the GOP size grows higher. In contrast, the proposed framework demonstrates better performance with more efficient compression rate in comparison with the quad tree algorithm. For instance, according to Figs. 6 and 7, with $M/N=0.1$ and GOP size of 5, the proposed algorithm sends 5 compressively sampled frames with the size of 0.1 of residual super-frame that is sent in quad-tree algorithm which gives us a gain of 2 in compression and saving the bandwidth up to 50 \% $(ii)$ Although the performance of the proposed framework outperforms that of the general video compression algorithms, it yields more computation complexity in the receiver. It is noteworthy that the complexity of the $\ell_1$-minimization method is of the order of $\mathcal{O}(N^{3})$, that of the OMP method is of the order of $\mathcal{O}(kMN)$ with $k$ as the order of sparsity, the one for CoSaMP method is of the order of $\mathcal{O}(\mathcal{L}\log(\|\mathbf{V}_{\sup }^{j}\|_{2}/\eta))$, where $\mathcal{L}$ is the multiplication cost of the sampling matrix, ${\Phi _{M \times N}}$, while $\eta$ is the precision parameter \cite{needell2009cosamp}. This is while for the quad tree algorithm, we have $\mathcal{O}((2p+1)QMN)$ as the order of complexity, and in comparison with OMP, there is a linear relation $kMN=h(2p+1)QMN$ with $h>1$ between their orders of complexity, where $p$ is the average search parameter for block matching algorithm, and $Q$ is the number of the quad tree blocks in each frame. By the way; the proposed method could be exploited for video streaming in WVSNs which have simple capturing nodes with low processing power and low bandwidth usage, but sink nodes with powerful processing power. It can also be used for UAV downlink video streaming in bandwidth restricted conditions. $(iii)$ More importantly, one of the great advantages of the proposed framework is the existence of the controlling factor, CT. Thereby, we can change the complexity of our framework to have a flexible performance-complexity trade-off. By increasing the CT we can suppress more non-zero elements, which leads to decreasing the iterations of inner-loops in sparse reconstruction algorithms, and also decreasing the total number of multiplications and additions which further confirming the merits of the proposed framework.
\section{Conclusion}
In this paper we proposed a framework for video streaming in reconnaissance and surveillance applications based on compressive sampling and sparse reconstruction for super-resolving residual super frames in scalable video sequences. The performance of the proposed framework evaluated in terms of compression rate, video quality and reconstruction complexity, and compared with recent state-of-the-art algorithms. The numerical results revealed that the proposed framework provides significant performance improvements in terms of compression rates and video qualities. Further, the flexibility of the proposed framework was provided by defining the CT as a controlling factor, and considering performance-complexity trade-off.
\begin{figure}
\centering
\includegraphics[width=9.5cm]{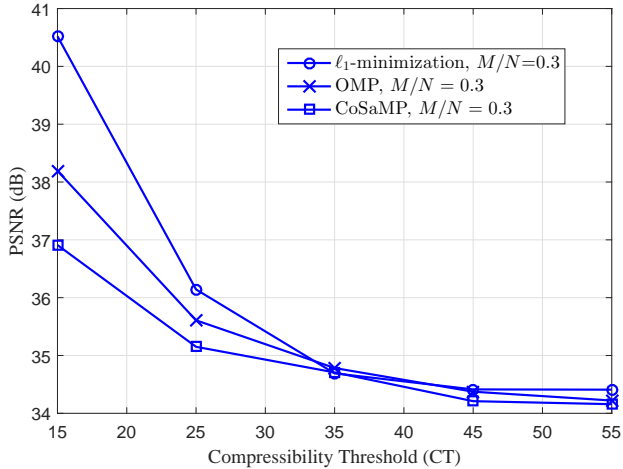}
\vspace{-1mm}
  \caption{Performance curves of $\ell_1$-minimization, OMP, and CoSaMP methods with $M/N$ = 0.3 for different values of CT.}
  \vspace{-5mm}
\label{}
\end{figure}

\begin{figure}
\centering
\includegraphics[width=9.5cm]{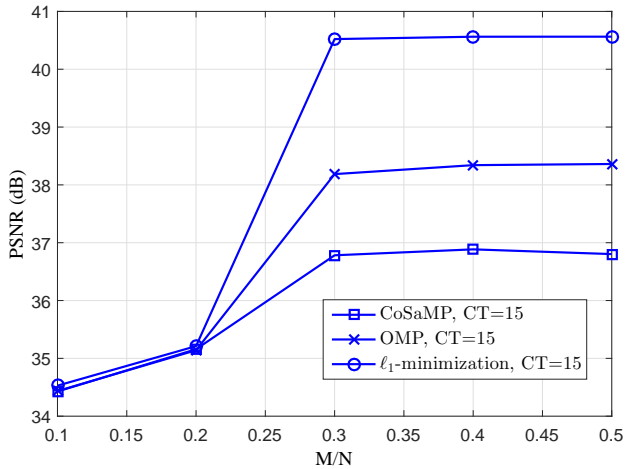}
\vspace{-4mm}
  \caption{Performance curves of $\ell_1$-minimization, OMP, and CoSaMP methods with CT = 15 for different values of $M/N$.}
  \vspace{-5mm}
\label{}
\end{figure}

\begin{figure}
\centering
\includegraphics[width=9.5cm]{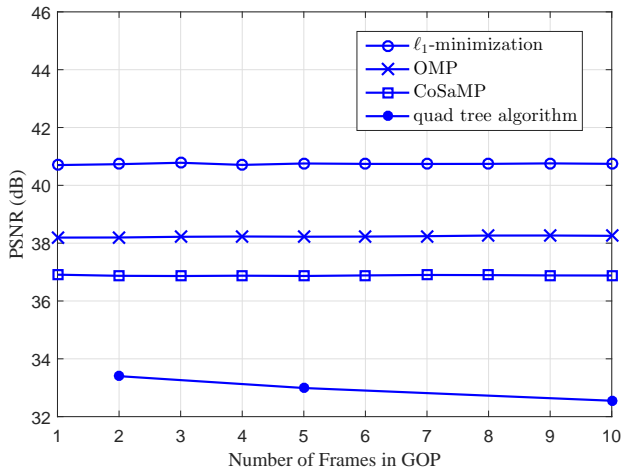}
\vspace{-4mm}
  \caption{Performance curves of $\ell_1$-minimization, OMP, and CoSaMP methods with $M/N$ = 0.3 and CT=15 according to proposed framework, and quad tree algorithm \cite{mahfoodh2015super} for first 10 frames of "in to the tree" video sequence.}
  \vspace{-5mm}
\label{}
\end{figure}

\small
\small

\bibliographystyle{IEEEtran}
\bibliography{IEEEabrv,refs_Mohammad}

\end{document}